\pdfoutput=1
\documentclass[11pt]{article}

\usepackage{acl}
\usepackage{times}
\usepackage{latexsym}

\usepackage[ruled]{algorithm2e}
\usepackage{booktabs}
\usepackage{times}
\usepackage{latexsym}
\usepackage{graphicx}
\usepackage{multirow}
\usepackage{multicol}
\usepackage{hyperref}
\usepackage{cite}
\usepackage{makecell}
\usepackage[T1]{fontenc}
\usepackage[utf8]{inputenc}
\usepackage{microtype}
\usepackage[flushleft]{threeparttable}
\usepackage{amsmath,amssymb,amsfonts}
\usepackage{scalerel,xparse}

\NewDocumentCommand\emojismile{}{
    {
        \includegraphics[scale=3]{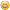}
    }
}
\NewDocumentCommand\emojicry{}{
    {   
        \includegraphics[scale=3]{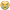}
    }
}

\title{TreeMix: Compositional Constituency-based Data Augmentation for Natural Language Understanding}

\author{Le Zhang \\
  Fudan University\\
  \texttt{zhangle18@fudan.edu.cn} \\\And
   Zichao Yang \\
   CMU\\ 
  \texttt{yangtze2301@gmail.com}\\\AND
   Diyi Yang \\ 
   Georgia Tech\\
  \texttt{dyang888@gatech.edu
} }

\date{}

\begin{document}
\maketitle

\begin{abstract}


Data augmentation is an effective approach to tackle over-fitting. Many previous works have proposed different data augmentations strategies for NLP, such as noise injection, word replacement,  back-translation etc. Though effective, they missed one important characteristic of language--compositionality, meaning of a complex expression is built from its sub-parts. Motivated by this, we propose a compositional data augmentation approach for natural language understanding called TreeMix. Specifically, TreeMix leverages constituency parsing tree to decompose sentences into constituent sub-structures and the Mixup data augmentation technique to recombine them to generate new sentences. Compared with previous approaches, TreeMix introduces greater diversity to the samples generated and encourages models to learn compositionality of NLP data. Extensive experiments on text classification and SCAN demonstrate that TreeMix outperforms current state-of-the-art data augmentation methods. We have publicly released our code \url{https://github.com/Magiccircuit/TreeMix}.
  \end{abstract}
\section{Introduction}
Data augmentation (DA) has won great popularity in natural language processing (NLP) \citep{chen2021empirical,feng2021survey} due to the increasing demand for data and the expensive cost for annotation. DA aims at increasing the quantity and diversity of the datasets by generating more samples based on existing ones, which helps make the training process more consistent and improves the model's capacity for generalization \citep{xie2020unsupervised}. 
For instance, existing DA methods often leverage word-level manipulation \citep{wei2019eda,kobayashi2018contextual,karimi2021aeda} and model-based sentence generation \citep{edunov2018understanding,ng2020ssmba}. As mixup-based \citep{zhang2018mixup} augmentation  achieving huge success in computer vision \citep{9008296,uddin2021saliencymix,kim2021comixup}, some recent works start to adapt mixup to NLP,  such as at the hidden level \citep{guo2019augmenting,chen2020mixtext} and at the input level \citep{yoon2021ssmix,shi2021substructure}.


\begin{table}[]\resizebox{7.7cm}{!}{
\begin{tabular}{|l|l|}
\hline
\multicolumn{2}{|l|}{\makecell{S1:They will find little interest in this poor film.}} \\ \hline
\multicolumn{2}{|l|}{\makecell{S2:It comes as a touching love story.}}   \\ \hline
\thead{Method\\}               & \thead{Example\\}                   \\ \hline
\makecell{EDA\\ \citep{wei2019eda}}         & \makecell{They will \textbf{this} find little\\ interest in \textbf{bad} \textbf{movie}.} \\ \hline
\makecell{AEDA\\\citep{karimi2021aeda}}         & \makecell{They will find \textbf{?} little in\\ \textbf{!} this poor movie\textbf{;}.} \\ \hline
\makecell{Noise\\\citep{xie2017data}}         & \makecell{\textbf{Thes} will \textbf{fi} little\\ \textbf{intres} \textbf{\_} this poor film .} \\ \hline
\makecell{SSMix\\\citep{yoon2021ssmix}}        & \makecell{They will find little interest\\ in  \textbf{love} poor film} \\ \hline
\makecell{Replacement\\\citep{kolomiyets2011model}}         & \makecell{They will find \textbf{limited} interest \\in this \textbf{odd} film.} \\ \hline
\makecell{Back Translation\\\citep{edunov2018understanding}}         & \makecell{They will \textbf{show} little interest \\in this \textbf{strange} film.} \\ \hline
\makecell{TreeMix\\}         & \makecell{They will find little interest \\in this \textbf{touching love story}.} \\ \hline

\end{tabular}}
\caption{Input-level DAs for Text-Classification. EDA includes random deletion, swapping, and insertion. AEDA randomly inserts punctuation. SSMix swaps tokens based on their saliency. The replacement method randomly substitutes words with synonyms.  In Back-translation, the source sentences are first translated into another language, and then back again.}
\label{tab:my-table}
\end{table}

Despite these empirical success, DA methods still suffer from key limitations. 
Simple rules based augmentation methods  \citep{wei2019eda,kobayashi2018contextual,karimi2021aeda} show little to none effect over large pretrained language models. 
While mixup-based augmentation methods demonstrate huge potential,
such interpolation at the hidden or input level has 
limited capability to capture explicit linguistic properties in text 
\citep{guo2019augmenting,chen2020mixtext,yoon2021ssmix}. 
Moreover, current DA methods exhibit limited ability in compositional generalization. Take a look at the following example from a BERT-based model that is fine-tuned using the SST2 dataset from the GLUE Benchmark:

\begin{table}[ht]
    \centering\small
    \begin{tabular}{lr}
    \hline
         This film is good and everyone loves it.&\emojismile 99\% \\ \hline
         This film is poor and I do not like it.&\emojicry 99\%  \\\hline
         This film is good and I do not like it.&\emojismile 99\%\\\hline
    \end{tabular}
\end{table}
\noindent
The first two examples are correctly classified. 
Despite that the last one is composed of fragments from the first two, the model fails to produce a correct or plausible label (in terms of characterizing a sentence's sentiment), demonstrating poor performance in compositional generalization.


However, compositionality is one key aspect of language that the meaning of a complex sentence is built from its subparts. 
Prior work also shows that 
syntax trees (e.g., tree-based LSTMs) are helpful to model sentence structures for better text classification \citep{Shi2018OnTN}.  
However, leveraging compositional structures for data augmentation has not received much attention in the language technologies communities, with a few exceptions in semantic parsing \citep{andreas2020goodenough,herzig2020span}. 

\begin{figure*}[htbp]
  \centering
  \includegraphics[width=1\textwidth]{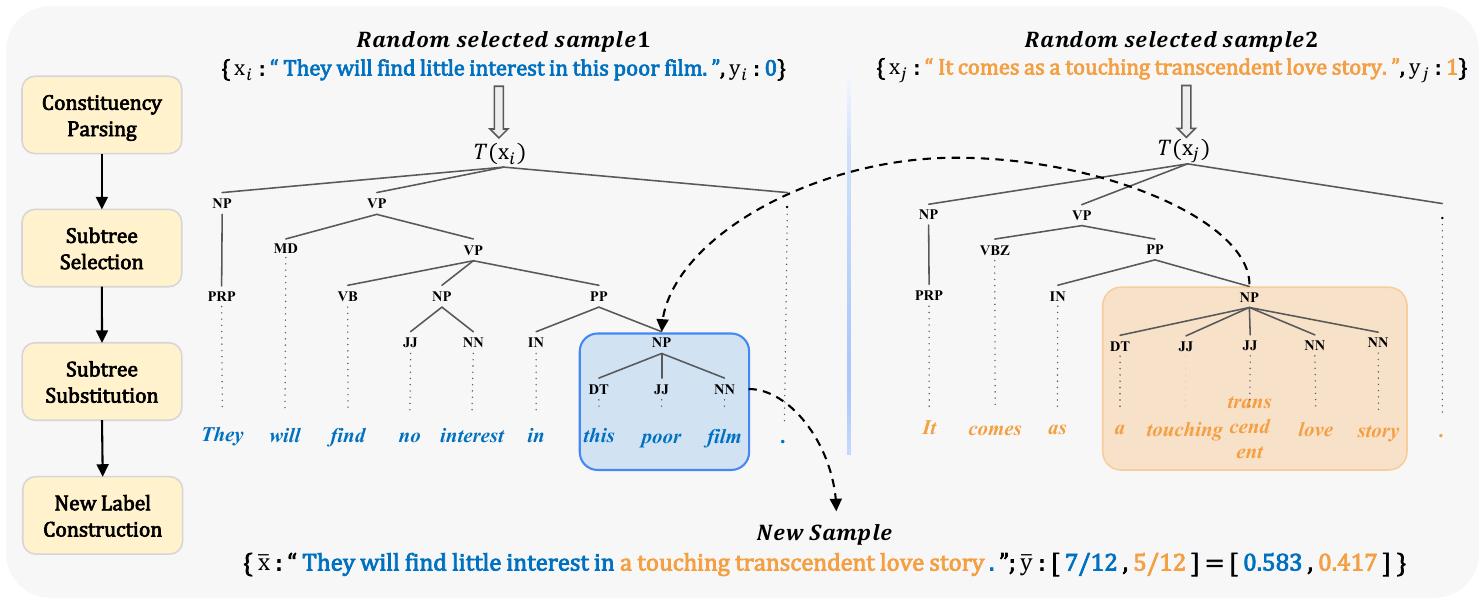}
  \caption{Illustration of TreeMix for single sentence classification}
  \label{fig:tree}
\end{figure*}

To this end, we propose a compositional data augmentation method for natural language understanding,  
i.e., TreeMix (Figure \ref{fig:tree}). TreeMix  is an input-level mixup method that utilizes constituency parsing information, where different fragments (phrase of a subtree) from different sentences are re-combined to create new examples that were never seen in the  training set; new soft labels will also be strategically created based on these fragments at the same time. In this way, TreeMix not only exploits compositional linguistic features to increase the diversity  of the augmentation, but also provides reasonable soft labels for these mixed examples.

Empirically, we find that TreeMix outperforms existing data augmentation methods significantly on a set of widely used text classification benchmarks. 
To validate the compositional effectiveness of TreeMix, we experiment with SCAN  \citep{lake2018generalization}---a task requires strong compositional generalization, and find that TreeMix exhibits reasonable ability to generalize to new structures built of components observed during training.

\section{Related Work}

\subsection{Generic Data Augmentation}
Most prior work operates data augmentation at different levels \citep{chen2021empirical}. 
\textbf{Token-level} DA methods manipulate tokens or phrases while preserving syntax and semantic meaning as well as labels of the original text, 
such as synonymy words substitutions \citep{Wang_2015,zhang2015characterlevel,Fadaee_2017,kobayashi2018contextual,Miao_2020} where synonyms are detected following pre-defined rules or by word embedding similarities. These methods has limited improvement \citep{chen2021empirical}  over large pretrained language models (PLMs). 
Besides, introducing noise by random insertion, replacement, deletion, and swapping \citep{wang2018switchout,wei2019eda,karimi2021aeda,xie2020unsupervised} is expected to improve the robustness of the model. 
\textbf{Sentence-Level} DA methods increase the diversity by generating distinct examples, such as via paraphrasing \citep{yu2018qanet,he2019revisiting,xie2020unsupervised,kumar2020data,chen2020mixtext,Cai_2020} or back translation \citep{Sennrich_2016,edunov2018understanding}. 
Other line of work used label-conditioned generation methods that train a conditional generation model such as GPT-2 or VAE to create new examples given labels as conditions \citep{Bergmanis_2017,Liu_2020a,Liu_2020b,Ding_2020,Anaby_Tavor_2020}.  
Although these methods can produce novel and diverse text patterns that do not exist in the original datasets, they require extensive training.  \textbf{Hidden-Level} DA methods mainly manipulate hidden representations by perturbation  \citep{Miyato_2019,Zhu2020FreeLBEA,Jiang2020SMARTRA,Chen2020SeqVATVA,Shen2020ASB,Hsu_2017,Hsu2018UnsupervisedAW,Wu2019DataAU,Malandrakis2019ControlledTG} and interpolation like mixup \citep{zhang2018mixup} to  generates new examples \citep{Miao_2020,Cheng2020AdvAugRA,chen2020mixtext,guo2019augmenting,guo2020sequencelevel,Chen2020LocalAB}.

\subsection{Compositional Data Augmentation}
Compositional augmentation aims at increasing the diversity of the datasets and improving the compositional generalization capability of the resulting models \citep{jia2016data,andreas2020goodenough}. 
These methods often recombine different components from different sentences to create new examples following a set of pre-designed linguistic rules such as lexical overlaps \citep{andreas2020goodenough}, neural-symbolic stack machines \citep{Chen2020CompositionalGV}, and substructure substitution \citep{shi2021substructure}. 
Compositional methods have been applied in a set of NLP tasks, such as sequence labeling \citep{guo2020sequencelevel}, semantic parsing \citep{andreas2020goodenough}, constituency parsing \citep{Shi2020OnTR,shi2021substructure}, dependency parsing \citep{Dehouck2020DataAV,shi2021substructure}, named entity recognition \citep{Dai2020AnAO}, text generation \citep{Feng_2020}, and text classification \citep{yoon2021ssmix,shi2021substructure}. Our work also falls into this category.

The most relevant  are \citet {shi2021substructure} and \citet{yoon2021ssmix}. However, \citet {shi2021substructure} 
only performs constituent substructure combinations with examples from the same category, thus inadequate in creating diverse enough augmentation with newly created labels.  

Besides, \citet{yoon2021ssmix} simply swaps the most and least salient spans, heavily relying on the model's performances in estimating salient spans,  and failing to consider these sentences' linguistic structures.
Our proposed TreeMix fills these gaps by allowing the composition of sentences from different label categories,  by utilizing rich consistency based structures in text, and by strategically generating soft labels for these augmented instances. 

\section{Method}
Our work is motivated by Mixup \citep{zhang2018mixup}, which creates virtual samples by mixing inputs. Given two random
drawn examples $(\mathbf{x}_i, \mathbf{y}_i)$ and $(\mathbf{x}_j, \mathbf{y}_j)$, where $\mathbf{x}$ denotes the 
input sample and $\mathbf{y}$ is the corresponding one-hot label, Mixup
creates a new sample by:
\begin{align}
    &\overline{\mathbf{x}} = \lambda \mathbf{x}_{i}+(1-\lambda) \mathbf{x}_{j},\nonumber\\
    &\overline{\mathbf{y}} = \lambda \mathbf{y}_{i}+(1-\lambda) \mathbf{y}_{j},\nonumber
\end{align}
where $\lambda \in [0,1]$. Mixup can be easily implemented in continuous space, 
hence some prior works~\citep{chen2020mixtext} have extended
it to NLP by performing interpolation in  hidden space. 

We improve upon Mixup by incorporating compositionality of language, a key characteristic that is essential
to generalization but neural models often fall short in capturing~\citep{lake2018generalization}. 
Instead of interpolating with the whole sample, TreeMix, our newly proposed method, creates new sentences by 
removing phrases of sentences and reinserting subparts from other sentences. TreeMix makes use of constituency
trees to decompose a sentence into meaningful constituent parts, which can then be removed and recombined to 
generate new augmentation samples. We aim to improve models' compositionality generalization ability by training
on large amount of samples produced by TreeMix. 
An example of using TreeMix for single sentence classification is shown in 
 Figure~\ref{fig:tree}.

\subsection{TreeMix}
\label{ssec:treemix}

Let $\mathbf{x}_i = \{x_i^1, x_i^2, ..., x_i^l\}$ denotes a sequence with length $l$ and its corresponding label
in one-hot encoding as $\mathbf{y}_i$. We run a constituency parser on $\mathbf{x}_i$ to get its parsing tree
as $T(\mathbf{x}_i)$. In order to get meaningful subparts of a sequence, we traverse the parsing tree recursively
and get all the subtrees with more than one child. Denote the collection of subtrees as 
$S(\mathbf{x}_i) = \{ t_i^k \}$, where $t_i^k$ denotes the $k$-th subtree of sample $\mathbf{x}_i$.
For a subtree $t_i^k$, it covers a continuous span $t_i^k \triangleq [x_i^{r_k}, ..., x_i^{s_k}]$ of $\mathbf{x}_i$
that starts with index $r_k$ and ends with index $s_k$. 
For example, as shown in the left part of Figure~\ref{fig:tree}, the subtrees of the example
sentence can cover spans such as  \texttt{this poor film},  \texttt{in this poor film}, 
\texttt{no interest} etc.

\begin{table}[]
\small
\resizebox{7.7cm}{!}{
\begin{tabular}{|ll|}
\hline
\multicolumn{2}{|c|}{They will find little interest in this poor film.}  \\ \hline
\multicolumn{1}{|l|}{{[}$\lambda_L,\lambda_U${]}}    & \multicolumn{1}{l|}{\textbf{possible selected sub-trees}}                                     \\ \hline
\multicolumn{1}{|l|}{{[}0.1,0.3{]}} & (little interest),(this poor film) \\ \hline
\multicolumn{1}{|l|}{{[}0.3,0.5{]}} & (in this poor film)                \\ \hline
\multicolumn{1}{|l|}{{[}0.5,0.7{]}} & (little interest in this poor film) \\\hline
\end{tabular}}
\caption{Examples of possible candidate subtrees with different $\lambda$ intervals}
\label{tab:subtree_example}
\end{table}
For a given sample $(\mathbf{x}_i, \mathbf{y}_i)$, we randomly sample another data point 
$(\mathbf{x}_j, \mathbf{y}_j)$ from the training set. We run the constituency parser on both sentences
and get their subtree sets $S(\mathbf{x}_i)$ and $S(\mathbf{x}_j)$, based on which we can sample
subtrees to exchange. We introduce two additional hyper-parameters $\lambda_L$ and $\lambda_U$ to constraint
the length of subtrees to sample. $\lambda_L$ and $\lambda_U$, measured in terms of length ratio of
the subtree to the original sentences, sets the lower and upper limits of the subtrees to sample. 
Intuitively, $\lambda$ controls the granularity of the phrases that we aim to exchange. 
We would
like that the length of phrase to exchange to be reasonable. If it is too short, then the exchange cannot introduce enough
diversity to the augmented sample; otherwise if it is too long, the process might inject too much noise to the original sentence. 
We set $\lambda$ to be the ratio in order to be invariant to the length of original sentences. 
Table \ref{tab:subtree_example} shows some subtree examples with different length constraints.
We define the length constrained subtree set as:
\begin{align*}
    S_\lambda(\mathbf{x}) \triangleq \{t | t \in S(\mathbf{x}), 
    s.t. \frac{|t|}{|\mathbf{x}|} \in [\lambda_L,\lambda_U]\}.
\end{align*}
Here $|.|$ denotes the length of a sequence or a subtree.
For two sentences $\mathbf{x}_i$ and $\mathbf{x}_j$, we randomly sample two subtrees
$t_i^k \in S_\lambda (\mathbf{x}_i)$ and $t_j^l \in S_\lambda (\mathbf{x}_j)$
and construct a new sample by replacing $t_i^k$ with $t_j^l$, i.e.
\begin{align}
    \bar{\mathbf{x}} \triangleq [x_i^1, ..., x_i^{r_k-1}, \underbrace{x_j^{r_l}, ..., x_j^{s_l}}_{t_j^l}, x_i^{s_k+1}, ... x_i^l]
\end{align}
where $t_j^l = [x_j^{r_l}, ..., x_j^{s_l}]$ replaces $t_i^k =[x_i^{r_k}, ..., x_i^{s_k}] $.
Figure.~\ref{fig:tree} shows an example of TreeMix, where the subtree 
 \texttt{a touching transcend love story} replaces the subtree  \texttt{this poor film}.

\begin{algorithm}[tbp]
  \caption{Dataset construction}
  \label{alg:algorithm1}
  \KwIn{Original dataset $\mathcal{D}$; data size multiplier $\beta$; parameters $\lambda_{L}$ and $\lambda_{U}$}
  \KwOut{Augmentation Dataset $\mathcal{D'}$}
  \While{$|\mathcal{D'}|$ \textless $\beta |\mathcal{D}|$} { 
  Randomly select two samples $(\mathbf{x}_i, \mathbf{y}_i)$ and $(\mathbf{x}_j, \mathbf{y}_j) \in {\mathcal D}$\\
  $(\bar{\mathbf{x}},\bar{\mathbf{y}})=\text{TreeMix}((\mathbf{x}_i, \mathbf{y}_i), (\mathbf{x}_j, \mathbf{y}_j))$
  $\mathcal{D'} \leftarrow \mathcal{D'} \cup \left\{(\bar{\mathbf{x}},\bar{\mathbf{y}})\right\}$}
\end{algorithm}

\paragraph{Label Creation for TreeMix} 
Creating a valid label for the augmented sample $\bar{\mathbf{x}}$ is a challenging problem. Similar
to that of Mixup \citep{zhang2018mixup}, we use a convex combination of original labels of two sentences as the new label
for the augmented sample:
\begin{align}\label{eq:two}
    \bar{\mathbf{y}} =\frac{l_i - |t_i^k|}{l_i - |t_i^k| + |t_j^l|} \mathbf{y}_i + \frac{|t_j^l| }{l_i - |t_i^k| + |t_j^l|}\mathbf{y}_j,
\end{align}
where $l_i$ is the length of $\mathbf{x}_i$ and $|t_i^k|, |t_j^k|$ are the length of the subtrees.
In the new sentence, $l_i - |t_i^k|$ words from $\mathbf{x}_i$ are kept and $|t_j^l|$ words from 
sentence $\mathbf{x}_j$ are inserted. 

In Equation~\ref{eq:two},  $\frac{l_i - |t_i^k|}{l_i - |t_i^k| + |t_j^l|}$
is the fraction of words that come from $\mathbf{x}_i$, which determines the weight of $\mathbf{y}_i$. 
The label is then created based on the conjecture that the change in labels is proportional to the length
changes in the original sentences.
We provided a set of augmentation examples from TreeMix in Table \ref{tab:addlabel} in Appendix. 


\paragraph{Pairwise Sentence Classification Task} 
The above mainly used single sentence classification as the running example for TreeMix. 
Here we argue that TreeMix can easily be extended to pairwise sentence classification problem, where
the relationship between the sentences is the label. 

Formally, for a given sample $(\mathbf{x}_i,\mathbf{x}'_i,\mathbf{y}_i)$, we randomly sample another sample $(\mathbf{x}_j,\mathbf{x}'_{j},\mathbf{y}_j)$ and run the parser and get the subtree sets of each sentence $S(\mathbf{x}_i),S(\mathbf{x}'_i)$ and $S(\mathbf{x}_j),S(\mathbf{x}'_j)$. Then we randomly sample subtrees $t_{i}^{k}\in S_{\lambda}(\mathbf{x}_i), t_{i'}^{k'}\in S_{\lambda}(\mathbf{x}'_i)$ and $t_{j}^{l}\in S_{\lambda}(\mathbf{x}_j), t_{j'}^{l'}\in S_{\lambda}(\mathbf{x}'_j)$. We construct $\bar{\mathbf{x}}$ by replacing $t_{i}^{k}$ with $t_{j}^{l}$ and $\bar{\mathbf{x}}'$ by replacing  $t_{i'}^{k'}$ with $t_{j'}^{l'}$. The new label is created as:
\begin{align}
    \bar{\mathbf{y}}=\frac{l_i+l_{i'}-|t_{i}^{k}|-|t_{i'}^{k'}|}{l_i+l_{i'}-|t_{i}^{k}|-|t_{i'}^{k'}|+|t_{j}^{l}|+|t_{j'}^{l'}|}\mathbf{y}_i+\\\nonumber
    \frac{|t_{j}^{l}|+|t_{j'}^{l'}|}{l_i+l_{i'}-|t_{i}^{k}|-|t_{i'}^{k'}|+|t_{j}^{l}|+|t_{j'}^{l'}|}\mathbf{y}_j.
\end{align}
The meanings of the notations are the same as in Equation~\ref{eq:two}.

Our main algorithm is shown in Algorithm \ref{alg:algorithm1}. Although not all sentences created by TreeMix 
are fluent or even valid new sentences, they contains subparts with different meanings that encourage
the models to build rich representation of sentences in a compositional manner. Note that the augmented labels
are convex combination of original labels, only when the model learns the representations of two parts 
together can they predict both labels with different weights.

\subsection{Training Objective}\label{ssec:merge_loss}



Our model is trained on a combination of the original samples and augmentation samples to obtain a trade-off
between regularization and noise injection. The final training objective is:
\begin{align}
    \mathcal L = & \mathop{\mathbb{E}}_{(\mathbf{x},\mathbf{y})\sim D}\left[ -\mathbf{y}^{\intercal} \log P_{\theta}(\mathbf{y}|\mathbf{x}) \right] \nonumber\\
     & + \gamma\mathop{\mathbb{E}}_{(\bar{\mathbf{x}},\bar{\mathbf{y}})\sim D'}\left[-\bar{\mathbf{y}}^{\intercal} \log P_{\theta}(\bar{\mathbf{y}}|\bar{\mathbf{x}})\right],
\end{align}
$\gamma$ is the weight\footnote{Section \ref{apdx:merge_loss} in Appendix presents discussions on how the objective and different weight parameter affects the result. } on the augmentation samples.

\section{Experiment}

\subsection{Datasets}
To test TreeMix's effectiveness, we experiment with a variety of text classification benchmarks, as shown in Table \ref{tab:dataset}.
We use accuracy as a metric, and exclude datasets from GLUE \citep{wang2018glue} that are not suitable for mixup, including CoLA that measures linguistic acceptability and will be ruined by mixup operations, and WNLI that is too small to show a method's validity.

\begin{table}[t]
  \centering
  \resizebox{7.7cm}{!}
    {\begin{tabular}{lccc}
      \bottomrule
      \multicolumn{1}{c}{Dataset} & Task       & Class & Size        \\
      \bottomrule
      \multicolumn{4}{c}{\textbf{Single Sentence Classification}}    \\
      \hline
      SST2                        & Sentiment  & 2     & 67k/1.8k    \\
      TREC-fine                  & Question   & 47    & 5.5k/500    \\
      TREC-coarse                & Question   & 6     & 5.5k/500    \\
      AG\_NEWS                    & News       & 4     & 12k/4k      \\
      IMDb                        & Sentiment  & 2     & 12.5k/12.5k \\\hline
      \multicolumn{4}{c}{\textbf{Pair Sentence Classification}}      \\ \hline
      RTE                         & NLI        & 2     & 3.5k/300    \\
      MRPC                        & Paraphrase & 2     & 3.7k/400    \\
      QNLI                        & QA         & 2     & 105k/5.5k   \\
      QQP                         & Paraphrase & 2     & 364k/40.4k  \\
      MNLI                        & NLI        & 3     & 393k/9.8k  \\
      \bottomrule
    \end{tabular}
    }
  \caption{Dataset name or split name, task category and number of label class, Size used for training and testing. For tasks from GLUE, Size indicates (\#train:\#validation); for TREC, AG\_NEWS and IMDb, Size indicates (\#train:\#test).   }
  \label{tab:dataset}
\end{table}

\subsection{Experiment Setup}
The proposed TreeMix method creates new samples by combining text spans based on the constituency tree's information, thus we use the Stanford CoreNLP toolkit\footnote{The specific version is 3.9.2} to obtain parsing related information \citep{manning2014stanford}. 
We use the pretrained language model \textit{bert-base-uncased} for sequence classification task from HuggingFace. With seeds ranging from 0 to 4 and $\lambda_{L}=0.1,\lambda_{U}=0.3$, we use TreeMix to generate twice and five times more samples than the original training set\footnote{Section \ref{sec:amountdata} in Appendix presents robustness check on how different amount of augmented data affects the result.}. We replicate the original dataset to the same size as the augmentation datasets in the training stage to ensure that the model receives the same amount of data from the original dataset and the augmentation dataset for each training batch.

If not specified, we train the model for 5 epochs, with a maximum sequence length of 128 and batch size of 96. The model is optimized using the AdamW optimizer with an eps of 1e-8 and a learning rate of 2e-5. 
Table \ref{tab:apd_set} in Appendix contains detailed hyper-parameter settings for each dataset.

\subsection{Baseline}
We compare TreeMix with the following benchmarks: 
(1) No augmentation (BERT): standard training without any augmentation, (2) EDA 
that randomly performs insertion, replacement, swap and deletion to the text. (3) AEDA that randomly inserts punctuation to the text. (4) Back translation(BT) \citep{edunov2018understanding}: texts are translated between English and German using Transformer architectures trained on WMT16 English-German. (5) GPT3Mix\citep{yoo2021gpt3mix} designs prompts and utilizes GPT3 to generate new examples to train the model. (6) SSMix \citep{yoon2021ssmix} applies mixup based on the saliency \citep{simonyan2014deep} of tokens, similar to PuzzleMix \citep{kim2020puzzle} and SaliencyMix \citep{uddin2021saliencymix}. (7) EmbedMix is the pretrained-language-model version of WordMixup in \citet{guo2019augmenting}, which performs mixup on the embedding level. (8) TMix \citep{chen2020mixtext} first encodes two inputs separately, then performs the linear interpolation of two embeddings at a certain encoder layer, and finally forward-passes the combined embedding in the remaining layers. 


\begin{table*}[t]
  \centering
  \begin{threeparttable}
    \resizebox{\textwidth}{!}{
      \begin{tabular}{lcccccccccc}
        \toprule
        \multicolumn{1}{c}{\multirow{2}{*}{Model}} & \multicolumn{5}{c}{Single Sentence Classification} & \multicolumn{5}{c}{Pair Sentence Classification}                                                                                                                                                                  \\
\cmidrule{2-11}                            & SST2                                               & TREC-f                                           & TREC-c                 & IMDb           & AG NEWS        & MRPC           & RTE            & QNLI           & QQP            & MNLI                            \\
\midrule
BERT                                       & 92.96\tnote{$\dagger$}                             & 92.36                                            & 97.08\tnote{$\dagger$} & 93.63          & 94.67          & 84.90          & 68.15          & 90.54          & 90.67          & 84.27\tnote{$\dagger$}          \\
BERT+EDA &92.20\tnote{$\dagger$}&91.95&96.79\tnote{$\dagger$} &93.62&94.67& -&-&-&-&-\\
BERT+AEDA &92.57\tnote{$\dagger$}&92.15&97.20\tnote{$\dagger$} &93.59&94.22& -&-&-&-&-\\
BERT+BT   &92.48&92.15&96.68 &-&-& 82.13&67.40&-&-&-\\           
BERT+GPT3Mix & 93.25\tnote{$\dagger$}&-&-&-&-&-&-&-&-&-                              \\
BERT+SSMix                                 & 93.14\tnote{$\dagger$}                             & 92.80                                            & 97.60\tnote{$\dagger$} & 93.74          & 94.64          & 84.31          & 68.40  & 90.60& 90.75   & \textbf{84.54\tnote{$\dagger$}} \\
BERT+EmbedMix  & 93.03\tnote{$\dagger$} & 92.32& 97.44\tnote{$\dagger$} & 93.72 & \textbf{94.72}  & 85.34          & 68.37          & 90.44          & 90.58          & 84.35\tnote{$\dagger$}          \\
BERT+TMix                                  & 93.03\tnote{$\dagger$}                             & 92.68                                            & 97.52\tnote{$\dagger$} & 93.69          & 94.69          & \textbf{85.69} & 68.45          & 90.48          & 90.66          & 84.30\tnote{$\dagger$}          \\\midrule
BERT+TreeMix                               & \textbf{93.92}                                     & \textbf{93.20}                                   & \textbf{97.95}         & \textbf{94.34} & \textbf{94.72} & 85.34          & \textbf{70.62} & \textbf{91.36} & \textbf{90.88} & 84.45                           \\
\bottomrule
      \end{tabular}%
    }
    \begin{tablenotes}
      \footnotesize
      \item[$\dagger$] denotes the result is extracted from the original paper
    \end{tablenotes}
    \caption{Results of comparison with baseline on full datasets, TREC-f and TREC-c indicates TREC-fine and TREC-coarse respectively. Scores are averaged over 5 random seeds. For GLUE tasks, we report accuracy of validation sets, and for other datasets we report test accuracy. EDA and AEDA will seriously damage the sentence relationship and harm the accuracy; GPT3Mix only reports full data experiments results on SST2 in original paper. We only report the results of back translation on small dataset due to the heavy computational cost.}
    \label{tab:result}
  \end{threeparttable}
\end{table*}

\section{Results and Analysis}

\subsection{Performance On Full Dataset} \label{ssec:full}

The results of TreeMix on the entire datasets are shown in Table \ref{tab:result}. TreeMix outperforms all baselines significantly on single sentence classification tasks, demonstrating the superiority of using compositional substructure for substitution and augmentation. For instance, On SST2, it improves by 0.98\%. Compared to other methods, the improvement was more than doubled. 

This is because that, unlike SSMix which substitutes the text spans based on the saliency, our TreeMix makes use of the constituency information to help identify linguistically informed sentence substructures, and by recombining these components, the compositional diversity of the datasets can be maximized. With our TreeMix generated samples, the model can see more combinations of the substructures in the training stage that aren't available in the original corpus, leading to better generalization ability. 

When it comes to sentence relationship classification tasks, TreeMix 
is also very effective. For example, It improves by 2.47\% on the RTE data set, whereas the best improvement of other methods is only 0.3\%, and it improves by 0.82\% on QNLI, where other data augmentation methods have little effect. We hypothesized that, when two constituent parts from one sentence pair are embedded into another sentence pair, the inherent relationship is also embedded. This better helps the models on how to  
to identify two pairs of relationships in a single sample, which further increases its capacity to categorize these challenging adversarial sentences.
Since TreeMix works by increasing dataset diversity and providing models with more text patterns to learn, it has very significant improvements
for these relatively small datasets such as RTE and TREC, compared to these large datasets such as AG NEWS,QQP and MNLI that already have a lot of diversity and text patterns.

\subsection{Influence of Constituency Information}\label{ssec:ci}

\begin{figure}[t]
  \centering
  \includegraphics[width=7.7cm]{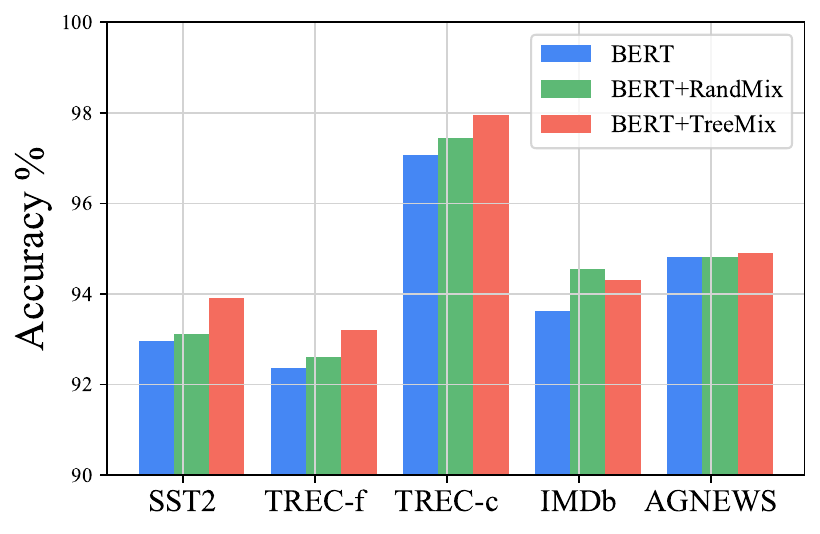}
  \caption{Performance of RandMix and TreeMix on single sentence classification datasets, scores are averaged over 5 random seeds.}
  \label{fig:rand_tree}
  \vspace{-0.2cm}
\end{figure}

To determine the importance of constituency information, we designed a  Random Mixup (RandMix) that randomly selects text spans as long as the ratio of span length to sentence length is less than a particular threshold $\lambda_{rand}$\footnote{We observed $\lambda_{rand} \sim \mathcal U(0,0.3)$ is optimal and we use this settings for the experiment}. The rest setting of RandMix is the same as TreeMix. We compare TreeMix and RandMix on single sentence classification datasets in Figure \ref{fig:rand_tree}.

We found that, both RandMix and TreeMix are quite effective, but TreeMix outperforms RandMix on most datasets. For instance, TreeMix exceeds RandMix by 0.8\% on SST2, 0.6\% on TREC-f, and 0.5\% on TREC-c.
One exception is on IMDb, where the average sentence length is much longer. The reason for the poorer performance of TreeMix is 
due to the sparse parsing results on long sentences; since 
there are many subtrees, substituting any single part might bring very minimal change to the entire sentence. 

\subsection{Influence of Training Set Size} \label{ssec:lr}
\begin{figure}[t]
  \centering
  \includegraphics[width=7.7cm]{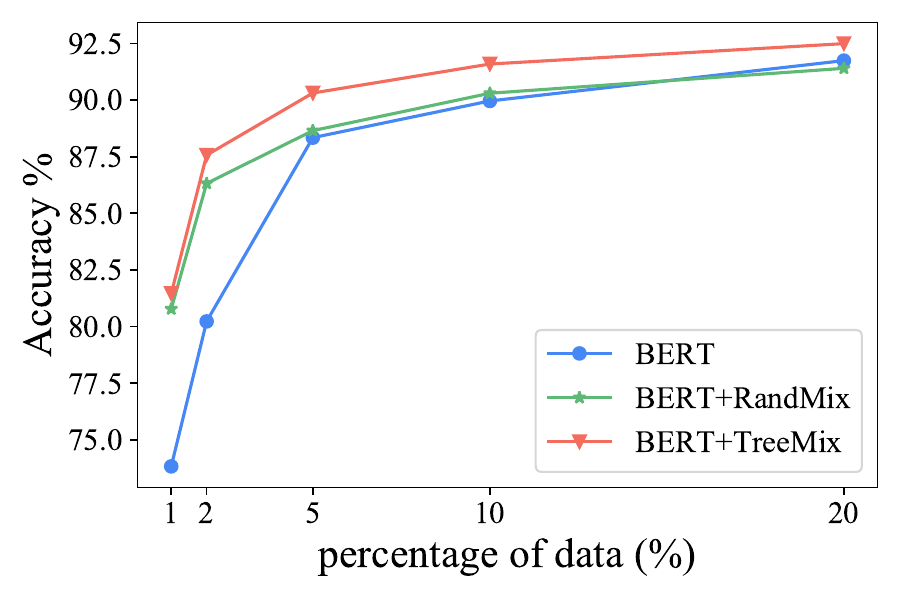}
  \caption{Results on SST2 varying data size. Scores are averaged over 5 random seeds.}
  \label{fig:sst2}
  \vspace{-0.2cm}
\end{figure}

To examine the influence of TreeMix with different training set sizes, we uniformly sample 1\%, 2\%, 5\%, 10\%, and 20\% of the data from the training set to investigate TreeMix in low-resource situations. 
The entire test set is used to evaluate the model's generalization ability. Since TreeMix generates more examples for training, we use RandMix to generate the same number of extra samples as a comparison to ensure the data size is fair. The results are summarized in Figure \ref{fig:sst2}. 

We found that, 
(1) TreeMix outperforms RandMix in all settings, further demonstrating the advantage of the compositional substructure with the constituency information over the randomly selected spans. (2) Both mixup methods can significantly improve the model's performance in the case of extreme data scarcity (e.g, 1\% and 2\%). 
(3) When the amount of data is sufficient (e.g, more than 5\%), TreeMix outperforms RandMix by a significant margin. However, TreeMix only slightly outperforms RandMix when there is a severe lack of data (e.g, 1\% and 2\%). This is due to that the too small datasets often contain very limited structures, thus constraining TreeMix's ability to increase text patterns and compositional diversity. (4) The relative improvement of TreeMix over conventional training without augmentation diminishes as the amount of data increases, largely due to 
that additional augmented text patterns might overlap with those already existing in the dataset, resulting in limited improvement.

\subsection{Influence of Cross-Category Mixing}
Different from prior work  \citet {shi2021substructure}, TreeMix allows the composition of sentences from different label categories. 
To test whether this cross-label category mixup is more effective than a within-label category mixup, we conducted ablation studies with TreeMix on samples in the same class\footnote{Section \ref{sec:samesubtree} in Appendix discuss the effect of constraining the length and label of the swapped subtree on the result}. 
Table \ref{tab:class} shows the results.
\begin{table}[t]
  \centering
  \resizebox{7.7cm}{!}{
  \begin{tabular}{lccc}
    \toprule
    Datasets & BERT  & TM(same)       & TM(cross)       \\
    \midrule
    SST2     & 92.96 & 93.78          & \textbf{93.92} \\
    TREC-fine   & 92.36 & 92.60          & \textbf{93.20} \\
    TREC-coarse   & 97.08 & 97.74          & \textbf{97.95} \\
    IMDb     & 93.63 & 94.22          & \textbf{94.34} \\
    AGNEWS   & 94.67 & 94.47          & \textbf{94.71} \\
    MRPC     & 84.90 & 85.34          & \textbf{85.34} \\
    RTE      & 68.15 & 70.25          & \textbf{70.62} \\
    QNLI     & 90.54 & 90.87          & \textbf{91.36} \\
    QQP      & 90.67 & 90.85          & \textbf{90.88} \\
    MNLI     & 84.27 & 84.33          & \textbf{84.45} \\
    \bottomrule
  \end{tabular}%
  }
  \caption{Performance with TreeMix performed (1) within same classes TM(same) and (2) cross different classes TM(cross), averaged over 5 runs.}
  \label{tab:class}%
\end{table}%
Across all datasets, we found that TreeMix that combines data from different classes is more effective than combining data from the same class, consistent with findings in \citet{zhang2018mixup}. When given only labels from one category, current models have a tendency to make simple or spurious judgments based on the most frequently occurring words. However the semantics of the sentence are complicated beyond simple words. For example, the model is likely to classify a sentence like \textit{``I like this good movie"} as positive because of the words \textit{ ``like"} and \textit{``good"}, but if \textit{``good movie"} is replaced with \textit{``bad film"}, the model must perceive the different constituent parts within the sentence. This ability can only be obtained  when the model is trained on the cross-category generated samples. 
\subsection{Influence of Length Ratio}\label{apdx:length}
\begin{table}[!htb]
  \centering
  \resizebox{7.7cm}{!}{
    \begin{tabular}{lcccc}
      \toprule
      Dataset     & \multicolumn{1}{l}{BERT} & \multicolumn{1}{l}{$\lambda=$[0.1,0.3]} & \multicolumn{1}{l}{$\lambda=$[0.3,0.5]}  \\
      \midrule
      SST2        & 92.96                    & \textbf{93.92}                        & 93.05                                 \\
      TREC-fine   & 92.36                    & \textbf{93.2}                         & 92.25                                 \\
      TREC-coarse & 97.08                    & \textbf{97.95}                        & 96.94                                 \\
      IMDb        & 93.63                    & \textbf{94.34}                        & 93.29                                 \\
      AG NEWS     & 94.67                    & \textbf{94.72}                        & 94.53                                 \\
      MRPC        & 84.90                    & \textbf{85.34}                        & 84.93                                 \\
      RTE         & 68.15                    & \textbf{70.62}                        & 70.35                                 \\
      QNLI        & 90.54                    & \textbf{91.36}                        & 90.78                                 \\
      QQP         & 90.67                    & \textbf{90.88}                        & 90.54                                 \\
      MNLI        & 84.27                    & \textbf{84.45}                        & 83.78                                 \\
      \bottomrule
    \end{tabular}
    }
  \caption{Performance with different length ratio intervals $\lambda$}
  \label{tab:lambda}
\end{table}

The only constraint we impose on TreeMix is the length ratio of the subtree controlled by $\lambda$. We select subtrees that are between 10\% and \%30 and between 30\% and 50\% of the length of the sentence, respectively. Table \ref{tab:lambda} shows the results.


On all datasets, $\lambda=[0.1,0.3]$ outperforms $\lambda=[0.3,0.5]$, which is in line with \citet{zhang2018mixup}'s observation  that giving too high mixup ration values can lead to underfitting. Another linguistic explanation for the scenario follows: When $\lambda=[0.3,0.5]$, TreeMix may select longer text spans, which usually contain unique constituency components like \textit{SBAR}; The exchange of these spans will severely damage the sentence's semantic and grammatical structure, causing the model to become confused. As a result, TreeMix with larger switching spans performs poorly, and even worse than baseline on some datasets.

\subsection{Compositional Generalization}
To quantify TreeMix's overall ability of compositional generalization beyond classification tasks, 
we conducted experiments on SCAN \citep{lake2018generalization} dataset, which is a command execution dataset widely used to test for systematic compositionality. It contains simple source commands and target action sequences. We test on commonly used challenging splits: \textit{addprim-jump, addprim-turn-left, around-right}, where primitive commands (e.g ``\emph{jump}'') only appear alone during training but will be combined with other modifiers (e.g ``\emph{jump twice}'') during testing. A model that works well for this task should learn to compose the primitive commands with the modifiers and generates corresponding execution. 
With TreeMix, we can generate the compositional commands that are not seen in the training set. 

The new command generation process is the same as in single sentence classification, except that we increase the length constraint $\lambda_U$ to 1 to allow the exchange of the commands with only one word. After we synthesize new commands, we follow the rules in \citet{lake2018generalization} to translate valid commands into actions and filter out ungrammatical commands. We follow the settings in  \citet{andreas2020goodenough} and use the following data augmentation methods as baselines: (1) WordDrop that drops words randomly; (2) SwitchOut \citep{wang2018switchout} that randomly replaces words with other random words from the same vocabulary;  (3) SeqMix \citep{guo2020sequencelevel} which creates new synthetic examples by softly  combining  in-put/output sequences from the training set, and (4) GECA \citep{andreas2020goodenough} that performs enumerated valid swaps.

As shown in Table \ref{tab:scan},  TreeMix outperforms SwitchOut and WordDrop for all splits. TreeMix by itself does not perform as well as GECA, but when being combined with GECA, it demonstrates very strong results. TreeMix outperforms SeqMix in all splits, due to the fact that TreeMix can more precisely find the linguistically rich compositional segments of a sentence, as evidenced by the results of the comparisons of TreeMix and SSMix in Section \ref{ssec:full} and TreeMix and RandMix in Section \ref{ssec:lr}.
\begin{table}[tb]
  \centering
  \begin{threeparttable}
      \resizebox{7.7cm}{!}{
        \begin{tabular}{lccc}
        \toprule
        \multicolumn{1}{l}{Method}      & \multicolumn{1}{l}{JUMP} & \multicolumn{1}{l}{TURN-L} &\multicolumn{1}{l}{AROUND-R} \\
        \midrule
        
        Baseline & 0\tnote{$\dagger$}    & 49\% \tnote{$\dagger$}&0\tnote{$\dagger$}\\
        WordDrop & 0\tnote{$\dagger$}    & 51\% \tnote{$\dagger$}&0\tnote{$\dagger$}\\
        SwitchOut & 0\tnote{$\dagger$}   & 16\% \tnote{$\dagger$}&0\tnote{$\dagger$}\\
        SeqMix & 49\%\tnote{$\dagger$}   & 99\% \tnote{$\dagger$}&0\tnote{$\dagger$}\\
        TreeMix & \textbf{72\%}    & \textbf{99\%} &\textbf{0\%}\\
        \midrule
        GECA  & 87\%\tnote{$\dagger$}    & -&82\tnote{$\dagger$}\\
        GECA+WordDrop & 51\%\tnote{$\dagger$} & -&61\tnote{$\dagger$}\\
        GECA+SwitchOut & 77\%\tnote{$\dagger$}&-&73\tnote{$\dagger$}\\
        GECA+SeqMix & 98\%\tnote{$\dagger$}    & -&89 \tnote{$\dagger$}\\
        GECA+TreeMix & \textbf{99\%}    & - &\textbf{91\%}\\
        \bottomrule
        \end{tabular}%
        }
        \begin{tablenotes}
          \footnotesize
          \item[$\dagger$] denotes the result is extracted from the original paper
        \end{tablenotes}
        \caption{Experimental results (accuracy) on SCAN. 
        }
      \label{tab:scan}%
  \end{threeparttable}
\end{table}%
A closer look at these augmented samples show that 
TreeMix can generate all possible combinations of ``\textit{jump}'' and other modifiers like ``\textit{left}'' and ``\textit{around}''; these previously unseen command combinations further validates TreeMix's ability to improve the dataset's compositional  diversity. 
TreeMix demonstrates weak performances on the \textit{around-right} split,  where the model observes commands ``\textit{around}'' and ``\textit{right}'' in isolation at the training stage, and it has to derive the meaning of ``\textit{around right}'' at the test time. Because the word ``\textit{around}'' cannot be parsed as a single subtree for swap. Instead, it always appears in a subtree with the word ``\textit{left}'', preventing TreeMix from generating the phrase ``\textit{turn right}''.
Despite its limitations on \textit{around-left}, TreeMix performs well on all other splits and can be easily combined with other data augmentation methods, 
demonstrating the compositional generalization ability of TreeMix beyond classification tasks.  


\section{Conclusion}
This work introduced TreeMix, a compositional data augmentation approach for natural language understanding. 
TreeMix leverages constituency parsing tree to decompose sentences into sub-structures and further use
the mixup data augmentation technique to recombine them to generate new augmented sentences. 
Experiments on text classification and semantic parsing benchmarks demonstrate that 
TreeMix outperforms prior strong baselines, especially in low-resource settings and compositional generalization. 

\section*{Acknowledgements} The authors would like to thank reviewers for their helpful insights and feedback. This work is funded in part by a grant from Salesforce. 

\bibliography{main}

\clearpage
\appendix

\numberwithin{table}{section}
\setcounter{page}{1}
\onecolumn
\section{Augmentation examples}
\begin{table}[htbp]
  \centering\small
    \begin{tabular}{|p{0.3\textwidth}|p{0.3\textwidth}|p{0.3\textwidth}|}
      \hline
      \multicolumn{1}{|c|}{Original Sentence1}& \multicolumn{1}{|c|}{Original Sentence2} & \multicolumn{1}{|c|}{New Sentence} \\
      \hline
      a love story and \textbf{a murder mystery} that expands into a meditation on the deep deceptions of innocence  [1] & really an advantage to invest such subtlety and warmth in an animatronic bear when the humans \textbf{are acting like puppets}  [0] & a love story and are acting like puppets that expands into a meditation on the deep deceptions of innocence [0.21 0.79] \\
      \hline
      the attempt to build up \textbf{a pressure cooker} of horrified awe  [0] & had the ability to mesmerize , astonish and entertain  [1] & the attempt to build up the ability of horrified awe [0.8 0.2] \\
      \hline
      rest contentedly with the knowledge that he 's made \textbf{at least one} damn fine horror movie  [1] & \textbf{minor film}  [0] & rest contentedly with the knowledge that he 's made minor film damn fine horror movie [0.13 0.87] \\
      \hline
      might just be better suited to \textbf{a night in the living} room than a night at the movies  [0] & are made for \textbf{each other} .  [1] & might just be better suited to each other room than a night at the movies [0.86 0.14] \\
      \hline
      is \textbf{a touching reflection} on aging , suffering and the prospect of death  [1] & keep upping the ante on \textbf{each other}  [1] & is each other on aging , suffering and the prospect of death [0 1] \\
      \hline
      is dark , brooding and slow , and takes \textbf{its central idea way}too seriously  [0] & \textbf{merely pretentious}  [0] & is dark , brooding and slow , and takes merely pretentious too seriously [1. 0.] \\
      \hline
  
    \end{tabular}%
    \caption{Examples of TreeMix on SST2 datasets, the number following sentence is label, bold tokens are selected phrase for substitution}
  \label{tab:addlabel}%
\end{table}%

\section{The necessity of merged loss techniques}\label{apdx:merge_loss}
We provide a detailed discussion of the techniques proposed in \ref{ssec:merge_loss}. We first investigate the noise contained in the augmentation dataset, then we figure out how the unbalance dataset will affect the performance. In the second part, we vary the weight parameter $\gamma$ to see how it affects the model's learning process. 
\subsection{Noise and Unbalance}
All mixup methods, as previously stated, introduce noise into the dataset. This noise in the text includes grammatical structure confusion and multiple semantic meanings in the sentences. The model will be overwhelmed by the noise if trained solely on the generated augmentation dataset, and will even perform worse than the baseline. In terms of the unbalance problem, we find that training the model without replicating the original dataset to the same size as the augmentation dataset hurts the model's performance. The results are shown in the table \ref{tab:noise}.

\begin{table}[htbp]
  \centering\small
    \begin{tabular}{llcccccccccc}
    \toprule
          &       & SST2  & TREC-f & TREC-c & IMDb  & AG NEWS & MRPC  & RTE   & QNLI  & QQP   & MNLI \\
    \midrule
    \multicolumn{2}{l}{BERT} & 92.96 & 92.36 & 97.08 & 93.63 & 94.67 & 84.9  & 68.15 & 90.54 & 90.67 & 84.27 \\
    \multicolumn{2}{l}{Merged Loss} & \textbf{93.92} & \textbf{93.2} & \textbf{97.95} & \textbf{94.34} & \textbf{94.72} & \textbf{85.34} & \textbf{70.63} & \textbf{91.36} & \textbf{90.88} & \textbf{84.45} \\
    \multicolumn{2}{l}{Augmentation only} & 92.57 & 90.44 & 96.42 & 92.37 & 93.98 & 83.93 & 65.45 & 88.72 & 89.24 & 83.78 \\
    \multicolumn{2}{l}{No Replicate} & 93.05 & 92.42 & 97.21 & 93.7  & 94.65 & 85.02 & 69.56 & 91.04 & 90.72 & 84.35 \\
    \bottomrule
    \end{tabular}%
    \caption{Merged Loss indicates results following techniques in \ref{ssec:merge_loss}, Augmentation indicates the model is trained on the generated dataset alone. No Replicate indicates Merged Loss without replication of the original training set.}
  \label{tab:noise}%
\end{table}%

\subsection{Weight parameter}
\begin{figure}[htbp]
  \centering
  \includegraphics[width=\textwidth]{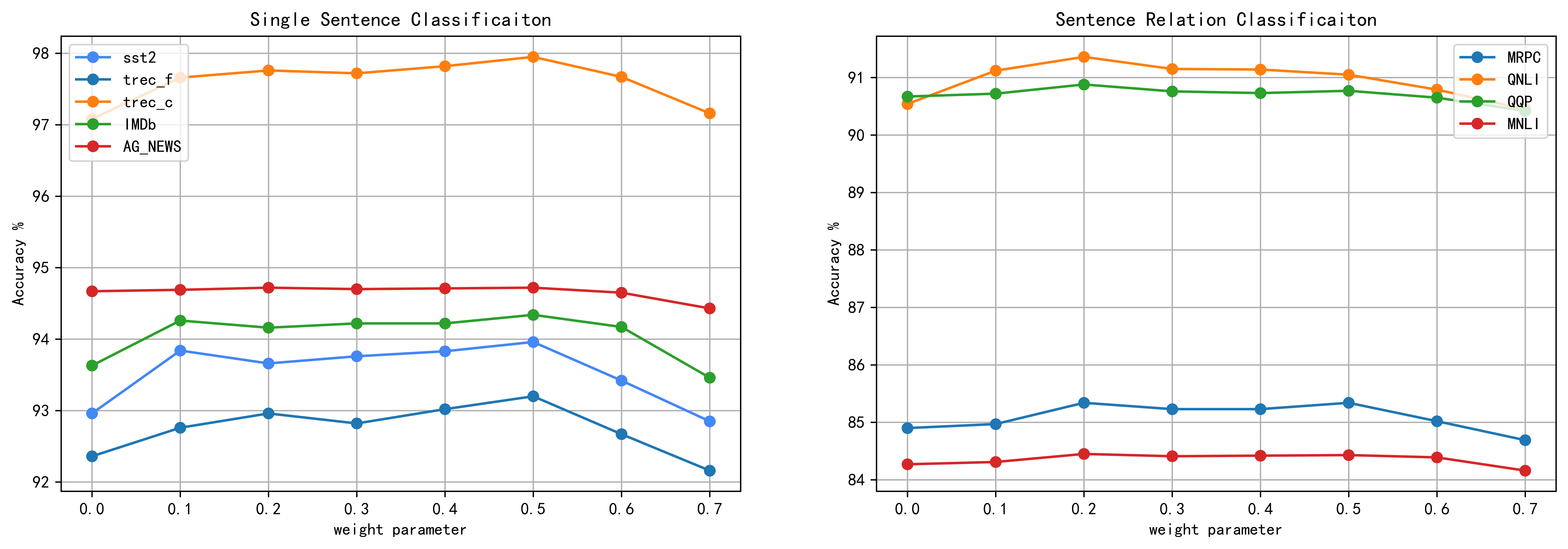}
  \caption{The performance when varying the value of the weight parameter on single sentence classification (left) and sentence relation classification (right)}
  \label{fig:weight}
\end{figure}

We vary weight parameter $\gamma$ to find optimal balance point between diversity and linguistic grammar, the results are shown in figure \ref{fig:weight}. Performance on the two classification tasks follows a similar pattern. Both increase with increasing weight and then rapidly decrease with increasing weight after reaching the highest point. Performance is weaker than the baseline when the weight value exceeds 0.7. We find the model achieves the best performance with $\gamma \in  \{0.2,0.5\}$. For single sentence classification tasks, when $\gamma=0.5$ the model always gets higher accuracy, and $\gamma=0.2$ is better for these sentence relation classification datasets.

\section{Hyper-parameters for each datasets}
\begin{table}[!htbp]
    \centering\small
      \begin{tabular}{lcccccc}
        \toprule
      Datasets & \multicolumn{1}{l}{epoch} & \multicolumn{1}{l}{batch size } & \multicolumn{1}{l}{aug batch size} & \multicolumn{1}{l}{val steps} & \multicolumn{1}{l}{sequence length} & \multicolumn{1}{l}{aug weight} \\
      \toprule
      SST2  & 5     & 96    & 96    & 100   & 128   & 0.5 \\
      TREC-f & 20    & 96    & 96    & 100   & 128   & 0.5 \\
      TREC-c & 20    & 96    & 96    & 100   & 128   & 0.5 \\
      IMDb  & 5     & 8     & 8     & 500   & 512   & 0.5 \\
      AGNEWS & 5     & 96    & 96    & 500   & 128   & 0.5 \\
      MRPC  & 10    & 32    & 32    & 100   & 128   & 0.2 \\
      RTE   & 5     & 32    & 32    & 50    & 128   & -0.2 \\
      QNLI  & 5     & 96    & 96    & 100   & 128   & 0.2 \\
      QQP   & 5     & 96    & 96    & 300   & 128   & 0.2 \\
      MNLI  & 5     & 96    & 96    & 100   & 128   & 0.2 \\
      \bottomrule
      \end{tabular}%
      \caption{Best settings for different datasets}
    \label{tab:apd_set}
  \end{table}

  We explore different parameter combinations and find the best ones for each task, as in Tab \ref{tab:apd_set}. There are some exceptions, such as TREC datasets, where the model cannot converge even with 10 epochs, so we increase the training epochs to 20 for this dataset. IMDb's examples are extremely long, with an average length of more than 200 words. Along with this change, we increased the truncation length to 512 and the batch size to 8 to fully capture the semantic meaning. RTE is the most unusual. First, when we train using original RTE datasets, the accuracy deviation is really substantial, reaching up to 4\%. Second, we find that $\gamma =-0.2$ is optimum for this set, which contradicts previous findings.
\section{Ablation Study}
\label{apdx:abs}
\citet{shi2021substructure} has proposed a similar study that uses constituency information for mixup. There are a few significant differences between our approaches. To begin with, their method is too restricted; they only perform mixup between examples from the same category, and they require the substituted subtree's label to be the same. Second, because they are limited to the same class examples, they are unable to devise a method for adding a soft label to the example. Instead, we only use TreeMix in the previous settings with the length constraint. Several other constraints in the subtree selection process are investigated in this section, and we achieve better performance than \citet{shi2021substructure} by giving the subtree selection process more freedom, and we validate that their work is a special case of our method by examining how other constraints affect the performance. This section's values are the averages of five runs with seeds ranging from 0 to 4

\subsection{What is the difference between different amounts of data?}\label{sec:amountdata}
TreeMix has the potential to generate an infinite amount of augmented data in theory. However, due to TreeMix's principle, it can only improve performance to a point when the size of the augmentation data set reaches a certain limit. We investigated how many augmentation datasets the model needs. Table \ref{tab:size} shows the results of producing twice and five times the augmentation data for experiments. 

\begin{table}[!htb]
  \centering\small
  \begin{tabular}{lllll}
    \toprule
    Dataset & Size  & BERT  & TM(x2)         & TM(x5)         \\
    \midrule
    RTE     & 3.5k  & 68.15 & 70.57          & \textbf{70.62} \\
    MRPC    & 3.7k  & 84.90 & 85.22          & \textbf{85.37} \\
    TREC-f  & 5.5k  & 92.36 & \textbf{93.2}  & 92.85          \\
    TREC-c  & 5.5k  & 97.08 & 97.71          & \textbf{97.95} \\
    IMDb    & 12.5k & 93.63 & \textbf{94.34} & 94.24          \\
    SST2    & 67k   & 92.96 & \textbf{93.92} & 93.92          \\
    QNLI    & 105k  & 90.54 & \textbf{91.36} & 91.34          \\
    AGNEWS  & 120k  & 94.67 & \textbf{94.71} & 94.69          \\
    QQP     & 364k  & 90.67 & \textbf{90.88} & 90.83          \\
    MNLI    & 393k  & 84.27 & \textbf{84.45} & 84.41          \\
    \toprule
  \end{tabular}
  \caption{Improvement of performance on all datasets with different amount of augmentation datasets, \textit{TM(x2)} indicates generating twice as much augmentation data than the original data, \textit{TM(x5)} indicates five times than original data, datasets in the table are listed in order of size}
  \label{tab:size}
\end{table}%

The key to getting the best results is to strike a balance between the original datasets and the augmentation datasets in terms of diversity and linguistic confusion. With more augmentation datasets, the model will learn more patterns while also observing more grammatically poor samples, which could negatively impact performance. We discovered that augmentation datasets twice the size of the original dataset produce the best results for larger datasets. This is in line with our previous theoretical analysis: large datasets inherently include more patterns and diversity, which helps the model generalize better. Maintaining the original linguistic grammar while increasing diversity in these datasets is, therefore, more important. When working with smaller datasets, it's better to train with more augmentation data. For models to train on these datasets, we believe diversity is more important than linguistic grammar.

TREC-fine is an exception. We attribute it to the datasets' excessive classes (up to 47 classes within only 5.5k training samples): each class has a very limited number of samples, and if we create overly augmented dataset samples, the limited samples of each category are insufficient to resist injected linguistic noise. As a result, for TREC-fine, x2 is preferable to x5. For a smaller dataset, we recommend generating five times as much augmentation data as possible, and for a larger dataset, we recommend generating twice as much augmentation data.

\subsection{Is it beneficial to keep the swapped subtree's label or length the same?}\label{sec:samesubtree}
\begin{figure}[htbp]
  \centering
  \includegraphics[width=0.5\textwidth]{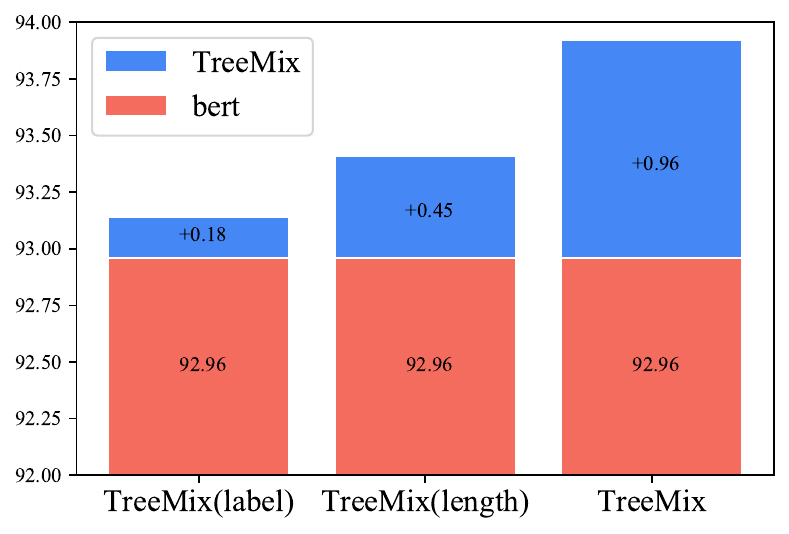}
  \caption{Performance on SST2 with different subtree selection constraints, green part is bert performance, orange part is improvement of TreeMix when applying different constraints, TreeMix(label) indicates only select subtrees with same phrase label, TreeMix(length) indicates only select subtrees with same length. TreeMix indicates without any constraints}
  \label{fig:sst2_phtase}
\end{figure}

Each subtree has its own label (e.g., VP and NP) and corresponds to a specific text span. When selecting subtrees, we can use these characteristics as additional constraints. Figure \ref{fig:sst2_phtase} shows the results. When we impose restrictions on the subtree selection process, the experimental results clearly show that performance suffers. 

We hypothesize that this is because in datasets with similar sentence lengths, subtrees of the same phrase label or phrase length tend to have similar structures (e.g., tree height, relative position in the sentence). Although the exchange of such subtrees can retain the original linguistic grammar of the text to some extent (e.g., replacing a noun phrase with another noun phrase will not significantly disrupt the sentence) and maintain similar sentence length, it cannot exploit the potential compositional diversity in the datasets as efficiently as TreeMix without any constraints, resulting in lower diversity augmentation datasets and limited improvement compared to the baseline.
In terms of the comparison of \textit{TreeMix(label)} and \textit{TreeMix(length)}, we find that \textit{TreeMix(label)} prefers simple phrases such as NP and VP because these are the most common phrases occurring in sentences, and this exchange will not improve the diversity of the datasets. For example, in "I like this apple," replacing "apple" with "orange" will not provide innovative text patterns.


\end{document}